\documentclass[letterpaper, 10 pt, conference]{ieeeconf}  

\IEEEoverridecommandlockouts                              

\usepackage{amsmath,amssymb,amsfonts}
\usepackage{bm}
\usepackage{graphicx}
\usepackage{textcomp}
\usepackage{xcolor}
\usepackage{enumerate}

\let\labelindent\relax
\usepackage{booktabs,siunitx,multirow,enumitem} 
\usepackage{subfigure}
\usepackage{hyperref}
\usepackage{cite}
\usepackage{color,soul}
\usepackage[ruled,norelsize,linesnumbered ]{algorithm2e}
\makeatletter
\newcommand{\removelatexerror}{\let\@latex@error\@gobble}
\makeatother
\SetCommentSty{mycommfontV}
\SetNlSty{}{}{}
\SetKwComment{SomeComment}{$\triangleright$\ }{}

\title{\LARGE \bf
GPU-Accelerated Policy Optimization via Batch Automatic Differentiation of Gaussian Processes for Real-World Control
}

\author{Abdolreza Taheri$^{1,4}$, Joni Pajarinen$^{2,3}$, and Reza Ghabcheloo$^{1}$
\thanks{*This work was funded by the European Union’s Horizon 2020 research and innovation programme Marie Skłodowska Curie grant No. 858101.}
\thanks{$^{1}$Faculty of Engineering and Natural Sciences, Tampere University, Finland. 
        {\tt\scriptsize \{reza.taheri, reza.ghabcheloo\}@tuni.fi}}%
\thanks{$^{2}$Department of Electrical Engineering and Automation,
Aalto University, Finland
        {\tt\scriptsize joni.pajarinen@aalto.fi}}%
\thanks{$^{3}$Intelligent Autonomous Systems, TU Darmstadt, Germany}%
\thanks{$^{4}$Control Systems R\&D, HIAB, Sweden}%
}

\begin{document}

\maketitle
\thispagestyle{empty}
\pagestyle{empty}

\begin{abstract}
The ability of Gaussian processes (GPs) to predict the behavior of dynamical systems as a more sample-efficient alternative to parametric models seems promising for real-world robotics research. However, the computational complexity of GPs has made policy search a highly time and memory consuming process that has not been able to scale to larger problems. In this work, we develop a policy optimization method by leveraging fast predictive sampling methods to process batches of trajectories in every forward pass, and compute gradient updates over policy parameters by automatic differentiation of Monte Carlo evaluations, all on GPU. We demonstrate the effectiveness of our approach in training policies on a set of reference-tracking control experiments with a heavy-duty machine. Benchmark results show a significant speedup over exact methods and showcase the scalability of our method to larger policy networks, longer horizons, and up to thousands of trajectories with a sublinear drop in speed.
\end{abstract}

\section{Introduction}
Recent advances in computational algorithms in Machine Learning, coupled with the increase in computing power have enabled researchers to take a modern approach to policy optimization. Reinforcement Learning (RL) is a popular method for optimizing policies for many complex tasks, by letting the algorithms interact with any evolving system and converge to a policy that generates the actions which maximize a cumulative reward for the task. Research in accelerating RL by leveraging multi-core architectures and hardware-accelerations provided by modern processors \cite{stooke2018accelerated,mnih2016asynchronous, schulman2017proximal,chatzilygeroudis2017black} has contributed significantly in shortening the training times.  

For a simulated environment, such as a dynamical system defined by a system of differential equations, an RL algorithm can effectively query the simulator many times, typically faster than it would take to do with a real system, and optimize the policy based on the collected trajectories \cite{langaaker2018cautious,andersson2021reinforcement,lindmark2018computational}. Simulated models often deviate from the real system dynamics, so the transferability of a policy that was trained in simulation to the real machine is a serious concern~\cite{muratore2021robot}. On the other hand, it is usually impractical and unsafe to run a real system for days or even months to collect training data to converge to a suitable solution. Among different techniques, model-based RL \cite{klink2021model,moerland2020model,janner2019trust,polydoros2017survey} offers a more feasible alternative, to build a surrogate model from the machine response, and optimize the control policies based on the obtained model. The advantage of this approach is clear: querying a fitted model is in many instances faster (and safer) than actuating the real system. Policy optimization works best if the model imitates closely the behavior of the real system, and since it is generally desired to learn while having as little interaction with the real machine as possible, many have used Gaussian Processes (GPs) \cite{williams2006gaussian} to predict the response of mechanical systems \cite{deisenroth2011pilco, deisenroth2014multi,calandra2014experimental,delgado2020sample,chatzilygeroudis2017black}. 

\begin{figure}[!tp]
  \centering
  {\includegraphics[width=0.95\columnwidth]{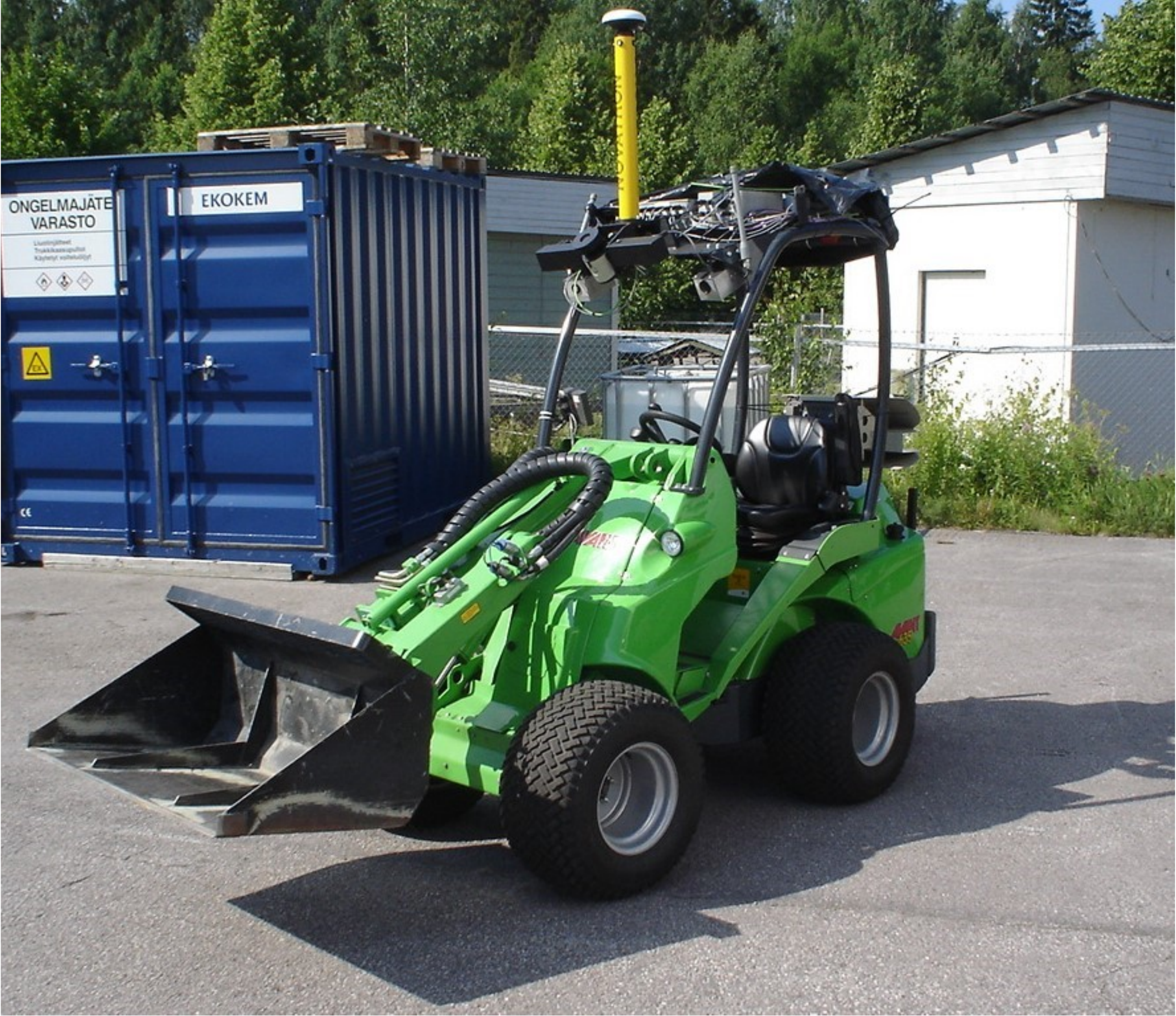}}
  \caption{The hydraulics of heavy-duty machines have complex dynamics which are difficult to model, and their lengthy maneuvers aggravate the size of dataset needed to predict their behavior. While previous methods with Gaussian processes crash for being out of memory or take hours to reach a lower performing single-goal solution (Section \ref{sec:experiment1}), our method BAGEL can train a policy in under 30 seconds on a laptop, which will drive the boom actuator of this machine accurately to any commanded position.}
  \label{fig:avant}
\end{figure}

Most RL algorithms that have utilized GPs in the past have been quite successful in solving robotics tasks \cite{deisenroth2013gaussian,chatzilygeroudis2017black}, having interacted with the real machine for only a limited time. While GP regression is a very effective method, the computational bottlenecks that accompany the previous inference methods (Section \ref{sec:relatedwork}) have made policy optimization a time and memory consuming process that scales worse with the size of training data. This has limited the research to simpler tasks, such as going from only one state to one goal using smaller datasets that can never extend beyond computational constraints. These challenges have to addressed for real-world applications (e.g. heavy-duty machines) where the maneuvers are slower and lots of training points constitute a well-descriptive model of the system for policy optimization. Fortunately, recent advances in GP inference methods have lifted the aforementioned constraints to a significant extent, providing faster and more efficient sampling (Section \ref{sec:relatedwork}).

\noindent\textbf{Contributions.} In this paper, we take a gradient-based optimization approach to obtaining a goal-conditioned nonlinear feedback policy for a dynamical system. In particular, we utilize GPs to represent a surrogate model of a heavy-duty machine. Then we use the model predictions to estimate the rewards and update the policy to maximize rewards. We build on top of previous methods in several directions: 
\begin{enumerate}
\item \textbf{Fast Inference.} At the core of our algorithm we utilize the fastest method for GP inference \cite{pleiss2018constant,gardner2018gpytorch} which is insensitive to the size of training data (Section \ref{sec:relatedwork}).
\item \textbf{GPU-Acceleration.} By mapping out the algorithm to process batches of trajectories in every pass (Section \ref{sec:policyeval}), we integrate vectorized Monte Carlo evaluations, GP inference, and automatic differentiation (autodiff) to run simulations and optimizations fully on GPU.   
\item \textbf{Flexibility.} We use the algorithm for training generalized  goal-conditioned policies in a timely fashion, and demonstrate control on a heavy-duty machine (Sections (\ref{sec:experiment1}-\ref{sec:experiment2gc})).   
\item \textbf{Scalability.} We provide benchmarks on the performance and learning speed of our algorithm and how it can scale to larger policies, longer horizon of predictions, and up to thousands of trajectories (Section \ref{sec:scalability}). 
\end{enumerate}
These provide us with a fast and practical framework for model-based optimization, accelerating learning while achieving good prediction accuracy that comes from the latest methods in GP inference. 

The structure of this paper is as follows: Section \ref{sec:relatedwork} overviews the related work and advances in algorithms, Section \ref{sec:methods} details our algorithm, Section \ref{sec:results} presents the results and benchmarks of our experiments and Section \ref{sec:conclusion} concludes the paper.  

\section{Related Work} \label{sec:relatedwork}

In real-world robotics applications, a key objective is to infer the policy with the least amount of interaction with the real machine, which is possible through imposing prior knowledge as well as incorporating models for system dynamics, rewards, and policy (for a detailed review of policy optimization in robotics, refer to \cite{chatzilygeroudis2019survey}). This is especially important in the field of heavy-duty machines, where dynamics are complex, interaction is slow and time-consuming, and tolerance is tight for unsafe exploration. Recently, there has been a surge in applying learning methods to heavy-duty machines \cite{backman2021continuous, kurinov2020automated,andersson2021reinforcement,lindmark2018computational,berglund2021controlling,yang2021neural} to discover automated, energy-efficient \cite{backas2019nonlinear} solutions for complex hydraulic systems. 

Gaussian Processes \cite{williams2006gaussian} are widely popular for modeling systems as accurately as possible where limited data exists. Having only a few hyperparameters, GPs are easy to tune, they provide the estimate of uncertainty over predictions, and can even be used to model rewards for a task \cite{chatzilygeroudis2017black}. Once a GP model is formed, it can be plugged into a planning algorithm to perform predictive control, i.e., to compute the sequence of actions which maximize the expected return over possible trajectories \cite{kamthe2018data,langaaker2018cautious}. A notable drawback in predictive control is the computational burden of online optimization which grows with the size of state and action spaces. An alternative approach is to optimize a parameterized policy offline \cite{deisenroth2011pilco, chatzilygeroudis2017black} and repeat the data collection and learning loop until the task is solved \cite{janner2019trust,klink2021model}. PILCO (probabilistic inference for learning control) \cite{deisenroth2011pilco} is one notable method among sample-efficient control approaches that uses GPs and explicitly propagates the uncertainty of the models. But doing so comes at the expense of significant $\mathcal{O}(n^3)$ complexity in time and $\mathcal{O}(n^2)$ in memory for $n$ training points. To speed up policy optimization, BLACK-DROPS \cite{chatzilygeroudis2017black} takes a black-box approach using the covariance matrix adaptation evolution strategy (CMA-ES), which enables the use of multiple cpu cores. Gradient-free optimization is a good approach for escaping local optima, but is in general slower on a single-core because it takes away a key feature, i.e. the gradient, which is in many instances computable. Since both PILCO and BLACK-DROPS use exact GP inference which grows cubically in computational complexity, and only PILCO can extend to GPU implementations, we choose PILCO as a baseline method in our benchmarks. 

Using exact GPs in model-based optimization does not imply error-free uncertainty propagation through the GP models. All works in learning with GPs use some form of approximation, in \cite{deisenroth2011pilco} the only tractable way to get the distribution of next states as a Gaussian is to approximate by matching moments. Additional approximations tackle the computational burden associated with the growing number of training points, for instance, sparse GP approximations \cite{snelson2007local, hewing2019cautious, snelson2006sparse}. In \cite{deisenroth2013gaussian} an approximate inference by linearizing the posterior mean of GP was shown to have less computational demand, yet slightly underperforming compared to moment matching. Pathwise conditioning \cite{wilson2021pathwise} is a recent approach that offers batch rollouts with sparse GPs, but is not studied in detail in the context of computation speed in RL. Moreover, all of these methods contribute to easing up the computations, but the degree of approximation and dependence on the size of training data are important factors that influence policy optimization on larger datasets and over longer-term predictions.


In spite of the computational constraints of GP inference that had existed for long, recent advances in algorithms have achieved significant speed ups with near-exact accuracy. The Kernel Interpolation for Scalable Structured GPs (KISS-GP) \cite{wilson2015kernel} is an inducing point method that precomputes a vector dependent on training data, allowing calculations of predictive mean by sparse interpolation in constant time (i.e., independent of number of training data). This was later extended to rapid computation of predictive covariances in constant time and sampling that is only linearly dependent on the number of test points in LanczOs Variance Estimates (LOVE) \cite{pleiss2018constant}, achieving thousands of times faster inference than previous methods. These are implemented through efficient matrix-vector multiplication (MVM) routines in GPyTorch \cite{gardner2018gpytorch}, which we will utilize for the predictive models. The advances in algorithms, coupled with memory efficient implementations of kernel operations \cite{charlier2021kernel,rudi2017falkon} and multi-GPU acceleration have enabled querying GPs up to millions \cite{wang2019exact} and billions \cite{meanti2020kernel} of training points. 

\section{Method} \label{sec:methods}
Our approach to policy optimization is based on the following objectives:
\begin{itemize}
\item Having limited interaction with the real system 
\item Solid convergence to high performing policies
\item Fast optimization, possibility to scale with hardware
\item Flexible implementation, generalized training
\end{itemize}
In this section, we will show how different elements come together in our algorithm to satisfy the above objectives. In brief, we will use a learned dynamics model from the machine response using Gaussian processes (Section \ref{sec:GPdynamics}) to optimize the policy based on the model as a proxy to the dynamics of the real system to address efficiency in interaction. To achieve the second objective, we propose the optimization objective as Monte Carlo evaluations over batches of trajectories for better estimate of gradient, and later confirm its effectiveness in the results of Section \ref{sec:experiment1}. The third objective is achieved by a)~using fast predictive GP sampling \cite{pleiss2018constant} (see Section \ref{sec:relatedwork}) and b)~unifying the algorithm to propagate and autodiff batches of states and actions (Section \ref{sec:policyeval}) to leverage GPU-acceleration, as benchmarked in Section \ref{sec:scalability}. Generalization to different problems is made possible through training over varying states and goals in batches, which can be sampled by any distribution, such that the policy maximizes the rewards over all states and goals.

\subsection{Probabilistic Dynamics Model using GPs} \label{sec:GPdynamics}

Model-based policy optimization \cite{klink2021model} is performed through approximating the dynamics using a model that is inferred from a dataset of transitions from the real system response. Gaussian processes (GPs)\cite{williams2006gaussian} are a class of probabilistic regression models that work for continuous state-action systems, and have been shown to be more sample efficient over many methods\cite{deisenroth2013gaussian,chatzilygeroudis2019survey}. GPs provide the amount of uncertainty in predictions, which can be incorporated into making policies more robust. Consider a system with states and actions $\bm{x} \in \mathbb{R}^p, \bm{u} \in \mathbb{R}^{q} $  whose dynamics is denoted by a transition
\begin{equation}
    \label{eqn_dynamics}
    \bm{x}_{k+1}=f_D(\bm{x}_k,{\bm{u}}_k)
  \end{equation}
A Gaussian process (GP) defines a distribution over functions $\hat{f}(\bm{x}) \sim \mathcal{GP}(\mu, k(\bm{x}_i,\bm{x}_j))$. Given a mean $\mu: \mathbb{R}^d \rightarrow \mathbb{R}$ and kernel function $k: \mathbb{R}^d \times \mathbb{R}^d \rightarrow \mathbb{R}$, each $m^\textrm{th}$-GP can model the one-step forward dynamics of one output $y^m_k = x^m_{k+1}$ \cite{umlauft2017learning} or the discrete difference $y^m_k = \Delta x^m_k = x^m_{k+1} - x^m_k$ \cite{deisenroth2011pilco}, of the states of the system $m=\{1,...,d\}$. To make notation simple, we will omit the superscript $m$ for each output function and overview the inference procedure which is similar for each GP. To make predictions using the GPs, the concatenated vector of states and actions $(\bm{x}_k,\bm{u}_k) \in \mathbb{R}^d$ is fed as training inputs and $\bm{y}_k$ as the corresponding training labels for all training points (subscripted with $k = \{1,...,n\}$) in a dataset of transitions $(X, \bm{y}) = \mathcal{D} \in \mathbb{R}^{n \times d}$. 
Given a test point $\bm{x}^\ast$, the predictive posterior $p(\hat{f}(\bm{x}^\ast)| X, \bm{y})$ is Gaussian distributed with mean and variance
\begin{align}
  &\mathbb{E}\left[ \hat{f}(\bm{x}^\ast)| X, \bm{y} \right] = \mu(\bm{x}^\ast) + \bm{k}^T_{X\bm{x}^\ast} \hat{K}^{-1}_{_{XX}} \bm{y} \label{eqn_GPmoments1} \\
  &\textrm{var} \left[ \hat{f}(\bm{x}^\ast)| X, \bm{y} \right] = k(\bm{x}^\ast, \bm{x}^\ast) - \bm{k}^T_{X\bm{x}^\ast} \hat{K}^{-1}_{_{XX}} \bm{k}_{X\bm{x}^\ast} \label{eqn_GPmoments2}
\end{align}
where $\hat{K}_{_{XX}} = (K_{_{XX}} + \sigma^2_n I)$, $K_{_{XX}}$ is the kernel matrix evaluated at all pairs, $\sigma^2_n$ is observation noise, and $\bm{k}_{Xx^\ast}$ is a vector of the kernel evaluated between test points $\bm{x}^\ast$ and the training points $X$. The squared exponential (SE) kernel with automatic relevance determination is one that is most frequently used in robotics
\begin{equation}
  \label{eqn_kernel}
  k(\bm{x}_i, \bm{x}_j)=\alpha^2 \textrm{exp}\left[-\frac{1}{2}(\bm{x}_i - \bm{x}_j)^T \bm{\Lambda} (\bm{x}_i - \bm{x}_j)  \right]
\end{equation}
where $\bm{\Lambda}$ is a diagonal matrix of length scales $\{l_1,...,l_{d}\}$ for each input dimension and $\alpha$ relates to function variance. The collective hyperparameters of the GPs $\bm{\phi} = [\bm{\Lambda}, \alpha, \sigma_n]$ are learned by maximizing the log marginal likelihood \cite{gardner2018gpytorch}
\begin{align}
  &\mathcal{L}_\mathcal{G} = \textrm{log}\, p(\bm{y} | X, \bm{\phi}) \propto - \bm{y}^T \hat{K}^{-1}_{\bm{\phi},_{XX}} \bm{y} - \textrm{log} |\hat{K}_{\bm{\phi},_{XX}}| \label{eqn_loglikelihood}\\
  &\frac{\partial \mathcal{L}_\mathcal{G}}{\partial \bm{\phi}} \propto  \bm{y}^T \hat{K}_{\bm{\phi},_{XX}} \frac{\partial \hat{K}^{-1}_{\bm{\phi},_{XX}}}{\partial \bm{\phi}}   \hat{K}_{\bm{\phi},_{XX}} \bm{y} -  \textrm{tr}\{\hat{K}_{\bm{\phi},_{XX}} \frac{\partial \hat{K}_{\bm{\phi},_{XX}}}{\partial \bm{\phi}}\} \label{eqn_loglikelihoodupdate}
\end{align}
In our algorithm, we use Lanczos variance estimates (LOVE) \cite{pleiss2018constant} implemented in GPyTorch \cite{gardner2018gpytorch} for sampling from the distribution of next states. Due to limited space, we refer to Section \ref{sec:relatedwork} for an overview, and to \cite{pleiss2018constant, gardner2018gpytorch} for details of these methods and implementation. Moreover, we focus on providing the benchmarks related to the performance of policy and optimization process that includes autodiff operations, whereas the speedups of these methods for inference-only have been documented in~\cite{pleiss2018constant,gardner2018gpytorch,wang2019exact}.

  \removelatexerror
   \begin{algorithm}[!tp]
  \small
   \SetAlgoLined
   \LinesNumbered
   \DontPrintSemicolon
    \textbf{Initialization} \\
    \Indp Parameterized policy $\pi_{\bm{\theta}}$, sample parameters according to \cite{he2015delving} or simply $\theta \sim  \mathcal{N}\left(0, \bm{I}\right)$\\
    Dataset of real transitions $\mathcal{D}$ - load or collect by applying actions manually or randomly\\
      Explicit reward function ${r}$ or (optional) collect a dataset $\mathcal{D}_R$ \\
  \Indm \textbf{for}  $n = 1 \rightarrow N_\textrm{interactions}$ \textbf{do} \\
      \Indp Learn GP transition dynamics using $\mathcal{D}$ (\ref{eqn_loglikelihoodupdate})  \\
      Learn GP reward model using $\mathcal{D}_R$ (Optional) \\
    \textbf{for} $i = 1 \rightarrow \textrm{max\_step}$ \textbf{until} convergance  \textbf{do} \\ 
     \Indp Sample batch of initial states $\mathcal{\hat{S}}^{b\times d}_0$ \SomeComment*[r]{Forward~pass} 
     Initialize return ${\hat{G}} \leftarrow {r}(\mathcal{\hat{S}}_0, \mathcal{G})$ \\
      \textbf{for} $k = 0 \rightarrow \textrm{Horizon-1}$ \textbf{do} \\
      \Indp ${U}_k \leftarrow \pi_{\bm{\theta}}(\mathcal{\hat{S}}_k, \mathcal{G})$\\
      Sample $\mathcal{\hat{S}}_{k+1} \sim \hat{f}_D(\mathcal{\hat{S}}_k, U_k)$ (\ref{eqn_GPmodel}) \\
      $\hat{G} = \hat{G} + {r}(\mathcal{\hat{S}}_{k+1}, \mathcal{G})$ \\
      \Indm \textbf{end for}\\
    Loss $\mathcal{L} \leftarrow -\frac{1}{b}[\textrm{Sum\_Reduce}(\hat{G})]$ \\
    Compute {$\nabla_\theta \mathcal{L}$} via Autodiff\SomeComment*[r]{Backward~pass} 
    Update $\bm{\theta}$ via gradient descent \\
    \Indm \textbf{end for}\\
    Collect more data (e.g. using $\pi_{\bm{\theta}}$) and update $\mathcal{D}$ ($\mathcal{D}_R$) \\
    \Indm \textbf{end for}
    \caption{Policy optimization via Batch Automatic differentiation of Gaussian process Evaluations using Lanczos variance estimates (BAGEL)}
    \label{Algo:I}
   \end{algorithm}

\subsection{Policy Evaluation \& Update} \label{sec:policyeval}
The RL objective for a system whose dynamics is defined by (\ref{eqn_dynamics}) is to find a policy $\pi_{\bm{\theta}}^\ast$ that maximizes the expected return of the closed-loop system response up to horizon H  
\begin{equation}
  \label{eqn_J}
  \mathbb{E}\left[ J(\bm{\theta})\right] = \sum_{k=0}^{H} r(\bm{x}_{_k}, \bm{x}_{g})
\end{equation}
where $r(\bm{x}, \bm{x}_g)$ provides an immediate reward for being in state $\bm{x}$ depending on the goal state $\bm{x}_g$. Throughout our experiments, we use the cost function used in \cite{deisenroth2011pilco} 
\begin{equation}
  \label{eqn_rew}
  r(\bm{x}, \bm{x}_g)=\textrm{exp}\left(-\frac{1}{2\sigma_r^2}(\bm{x} - \bm{x}_g)^T \bm{Q} (\bm{x} - \bm{x}_g)  \right)
\end{equation}
where $\bm{Q}$ and $\sigma_r$ are weighting and width-adjusting parameters. At every forward pass, trajectories are realized according to the actions provided by the policy $\bm{u}_k = \pi_{\bm{\theta}} (\bm{x}_k, {\bm{x}_g})$ and GP models 
\begin{equation}
  \label{eqn_GPmodel}
  \hat{f}_D(\bm{x}_k, \bm{u}_k) \sim \mathcal{N}(\bm{\mu}(x^\ast_k), \bm{\Sigma}(x^\ast_k))
\end{equation}
with the moments defined in (\ref{eqn_GPmoments1})-(\ref{eqn_GPmoments2}). In our experiments we consider feedforward Neural Network (NN) policies with deterministic outputs and exploit the gradient of policy parameters that become available when system transitions are in GP form with parameterized rewards and policies \cite{deisenroth2011pilco}. Instead of analytically derived gradients, we use autodiff in PyTorch \cite{paszke2019pytorch} for a straightforward and flexible implementation. To obtain a good estimate of the objective function (and the gradients), Monte Carlo sampling requires many trajectories to be unrolled. Therefore, the objective function in a forward pass over $b$ trajectories is obtained by
\begin{equation}
  \label{eqn_G}
  \mathbb{E}\left[ G(\bm{\theta})\right] = \frac{1}{b}\sum_{\tau=0}^{b} \sum_{k=1}^{H} \hat{r}(\hat{\bm{x}}_{_{k-1}} + \hat{f}_D(\hat{\bm{x}}_{k-1}, \bm{u}_{k-1}) , \bm{x}_{g})
\end{equation}
where $\hat{\bm{x}}$ denotes the predicted state using the surrogate models. Furthermore, we reformulate the objective function to exploit batch querying GPs for the distribution of next states
\begin{equation}
  \label{eqn_GB}
  \mathbb{E}\left[ G_{_B}(\bm{\theta})\right] = \frac{1}{b}\sum_{k=1}^{H} \hat{r}(\hat{\mathcal{S}}_{_{k-1}} + \hat{f}_D(\hat{\mathcal{S}}_{k-1}, \pi_{\bm{\theta}} (\hat{\mathcal{S}}_{k-1},\mathcal{G})) , \mathcal{G})
\end{equation}
by passing states and goals in batches $\{\mathcal{S}, \mathcal{G}\}$. Finally, the gradient of objective function $\nabla_\theta \mathcal{L}$ can be computed by reducing (\ref{eqn_GB}) by summation and autodiff to update the policy parameters, for which we will use Adam \cite{kingma2014adam}. Algorithm~\ref{Algo:I} summarizes our policy optimization procedure termed BAGEL. This implementation is simple yet flexible for different formulations. For instance, the initial and goal states $\{\mathcal{S}, \mathcal{G}\}$ at every iteration can be fixed for training as in \cite{deisenroth2011pilco,chatzilygeroudis2017black} which we will demonstrate in our first experiment in Section \ref{sec:experiment1}, or they can be sampled according to a distribution (Section \ref{sec:experiment2gc}) for a generalized policy. More importantly, all operations in Algorithm~\ref{Algo:I} can be performed on GPU, for which we will provide detailed speedups over CPU (Section \ref{sec:experiment1}) and scalability to thousands of trajectories (Section \ref{sec:scalability}).

\section{Results \& Discussion} \label{sec:results}
\subsection{Experiment 1: Single-goal Controller} \label{sec:experiment1}
\begin{figure}[!tp]
  \centering
  \subfigure[]{\includegraphics[width=0.95\columnwidth]{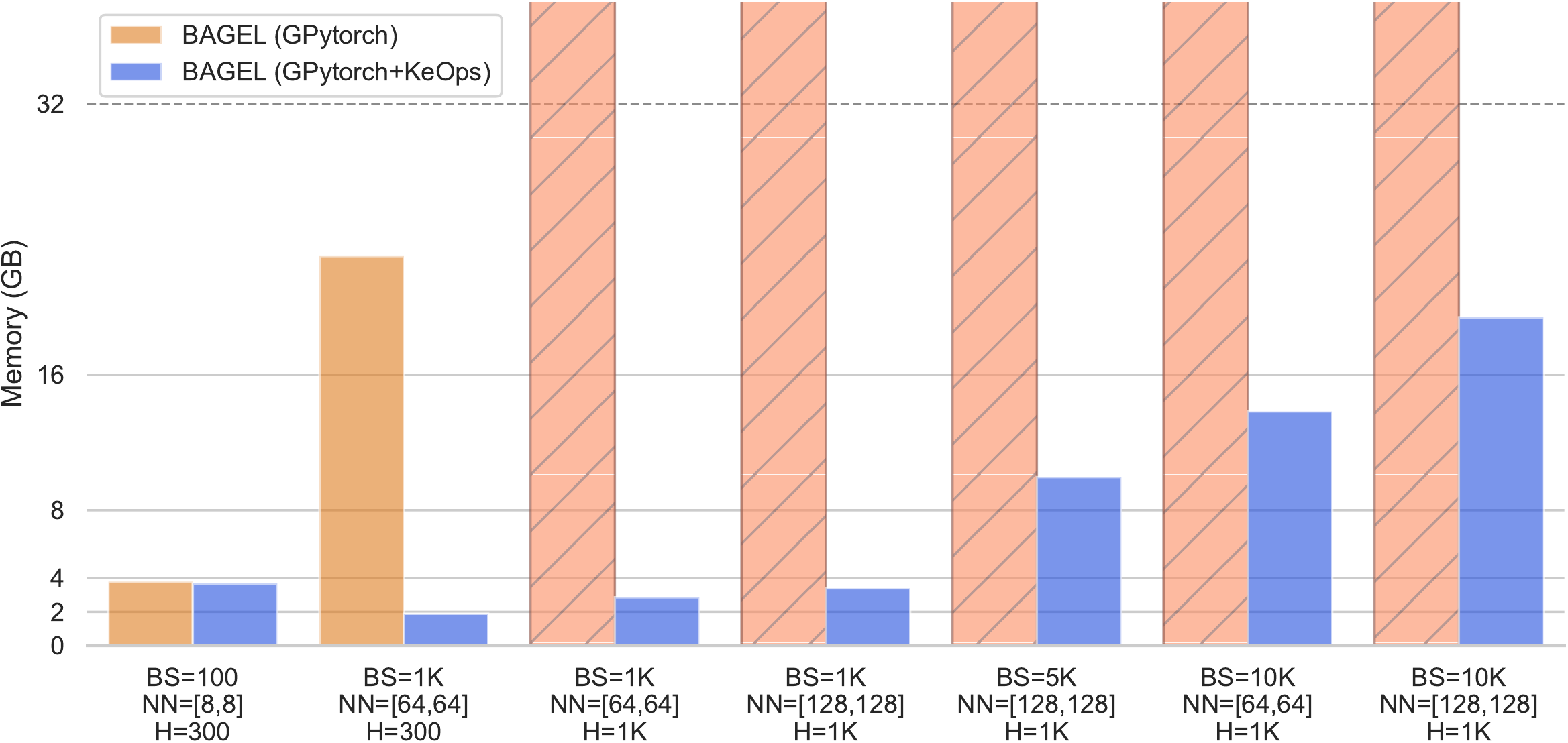}} 
  \subfigure[]{\includegraphics[width=0.95\columnwidth]{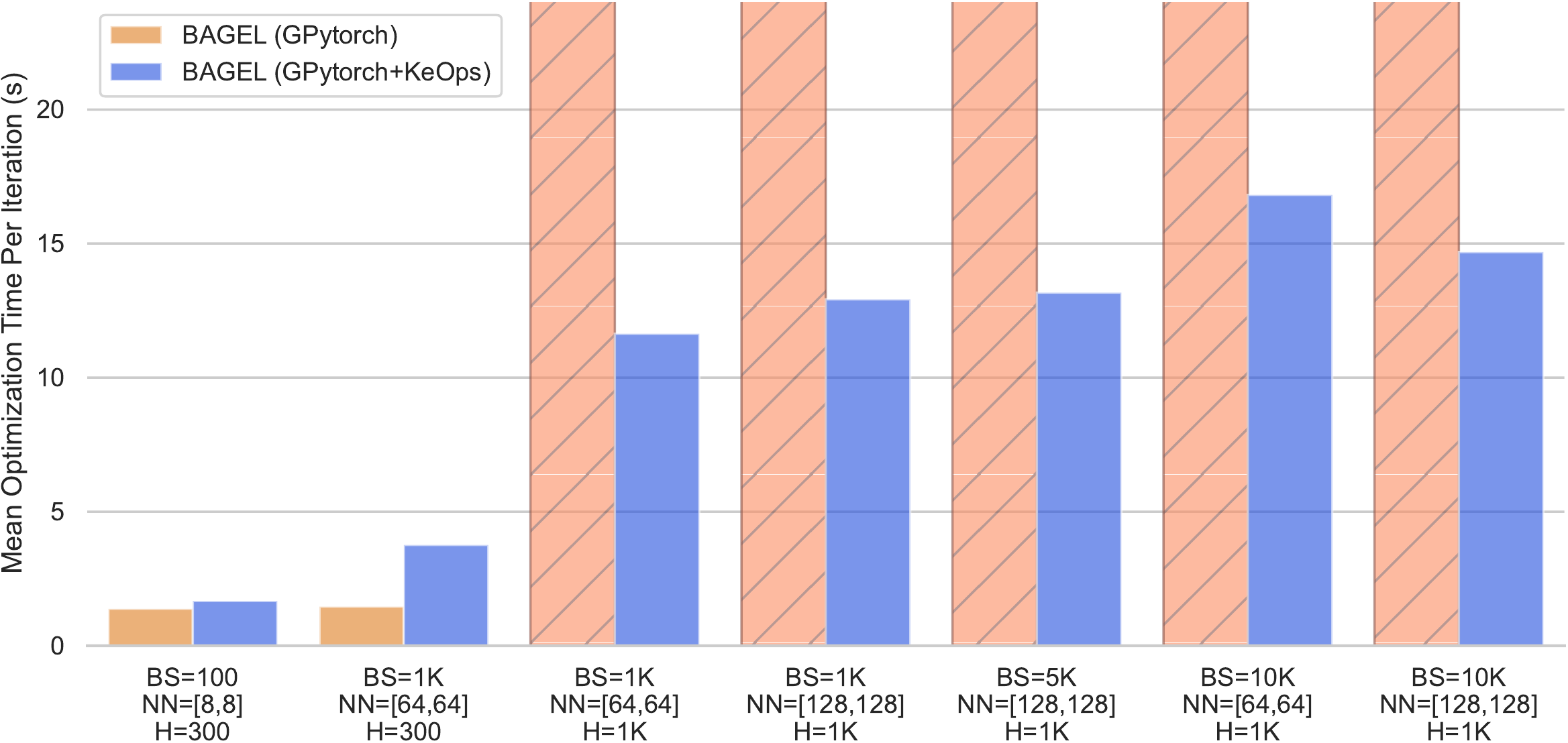}} 
  \caption{Benchmarks of (a) memory and (b) speed of BAGEL implementations for different configurations. BS: Batch Size, H: horizon of forward pass, NN: Neural Network policy layers. Hatched bars mark configurations that were unable to run on GPyTorch-only implementation due to out of memory error. The benchmarks use a NVIDIA Tesla V100 GPU from CSC IT Center for Science Finland.}
  \label{fig:keops}
\end{figure}

In our first experiment, we formulate the task as driving the hydraulic boom of the machine depicted in Fig.~\ref{fig:avant} from a low angle to the horizontal pose (i.e. $0^\circ$). To keep the formulation simple and provide a comparison between BAGEL and exact methods, we consider the kernel in (\ref{eqn_kernel}) for the GP models. We use a batch size of 100 trajectories, horizon of 300 steps and choose two states for GPs: the angle $\varphi$ and angular rate $\dot{\varphi}$ of the hydraulic boom, which are noisy observations from the real machine. The batch of initial states $\hat{\mathcal{S}} \in \mathbb{R}^{100 \times 2}$ is initialized with all repeating entries of the current state of the machine i.e. $\hat{\mathcal{S}}_k = {[\hat{\varphi}_0, \hat{\dot{\varphi}}_0]}$, normalized according to the mean $\bm{\mu}_X$ and standard deviation $\bm{\sigma}_X$ of the data in $\mathcal{D}$. For the reward function, we use $r([\hat{\varphi}, \hat{\dot{\varphi}}], \frac{-\bm{\mu_X}}{\bm{\sigma_X}})$ in (\ref{eqn_rew}) with $\bm{Q} = \textrm{diag}\{10,0.1\}$ and for the policy we use a NN with two hidden layers of 8 nodes each (referred to as [8,8] in Fig.~\ref{fig:keops}) whose outputs are bounded in the range [-1,1] using a saturating function.

\begin{figure}[!tp]
  \centering
  {\includegraphics[width=0.99\columnwidth]{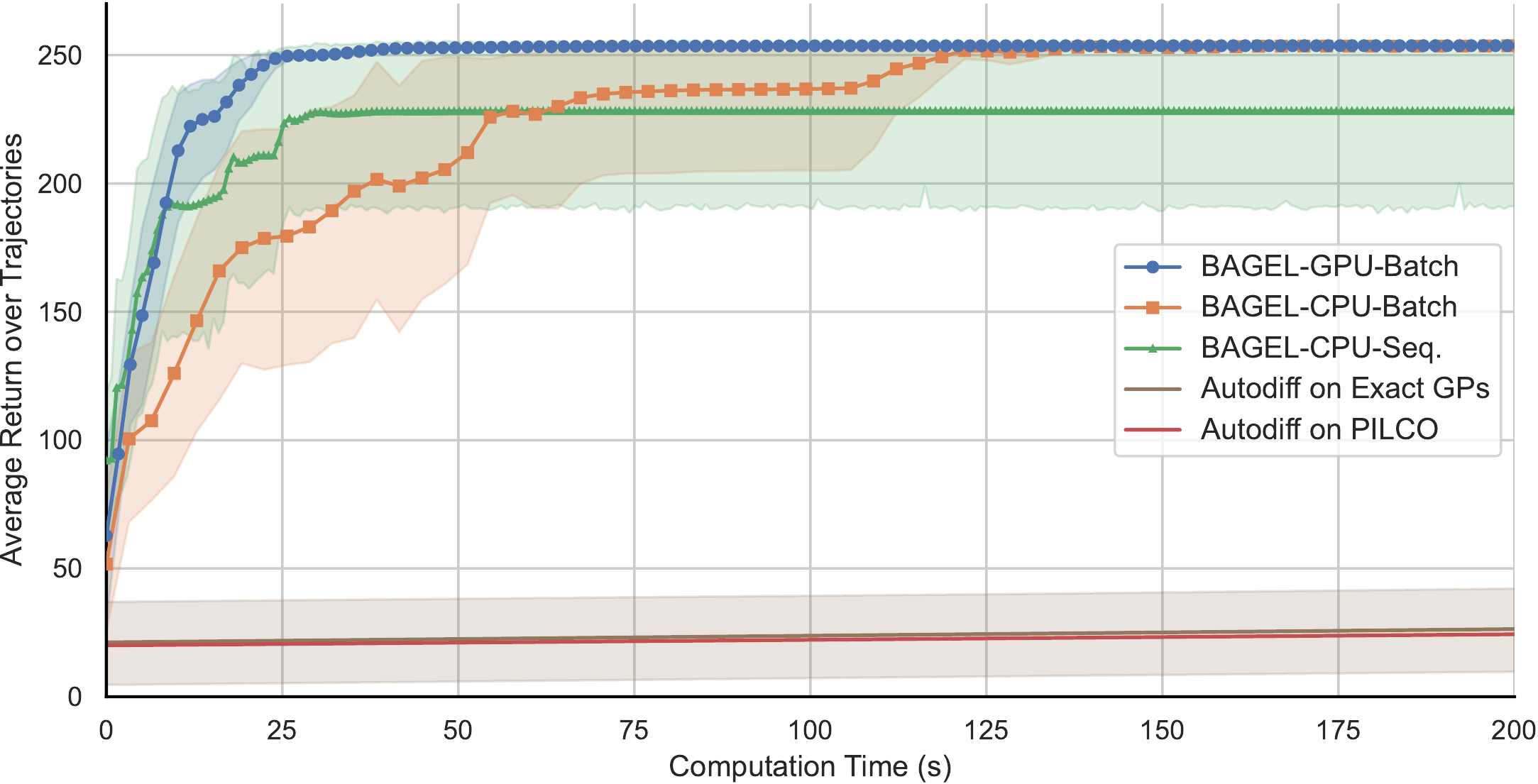}}
  \caption{Policy optimization progress across GPU- and CPU-accelerated BAGEL versus other methods. Batch implementations process 100 trajectories on every forward pass, \textit{Seq} refers to sequentially querying the GPs, i.e. one trajectory in every pass. Mean and 95\% confidence bounds over 16 trials are shown for each method. In this experiment with $n=2200$ training data, exact methods frequently crash due to memory overflow. Computations are made using a laptop Intel 10750H CPU and NVIDIA Quadro T2000 GPU.}
  \label{fig:ComputeTime}
\end{figure}

To provide a good comparison between algorithms, we train the GPs on a dataset $\mathcal{D}$ of previously collected transitions from manually lowering and raising the boom for 110~seconds. With a sampling rate of $0.05s$, there are $n=2200$ training points in $\mathcal{D}$. We use the same GPs for all algorithms and do not iterate for collecting more data, as we consider the GPs to be sufficiently descriptive of the dynamics and are only interested in comparing the optimization performance under similar circumstances. Fig.~\ref{fig:ComputeTime} summarizes the optimization time of different methods. As can be seen in the results, both the GPU and CPU versions of BAGEL with batch-size=100 achieve the highest reward in all trials, with GPU converging 5x faster than CPU since CPUs fall short in batch processing, motivating the need for GPU-acceleration. To support the effectiveness of batch evaluation in BAGEL, optimization is also shown over a single trajectory per pass under CPU-Seq, as if batch-size=1. Here, the single CPU core can perform sequential updates much faster than a GPU (each update illustrated with a marker), achieving a very high rate of updates yet falling short in achieving better quality of updates over GPU-Batch. This is because one rollout is not fully descriptive of the uncertainty of the models, the estimate of gradient is noisy and the optimization can eventually get stuck in local optima. The learning rate used for all trainings is $10^{-2}$.

\begin{figure}[!tp] 
  \centering
  {\includegraphics[width=0.99\columnwidth]{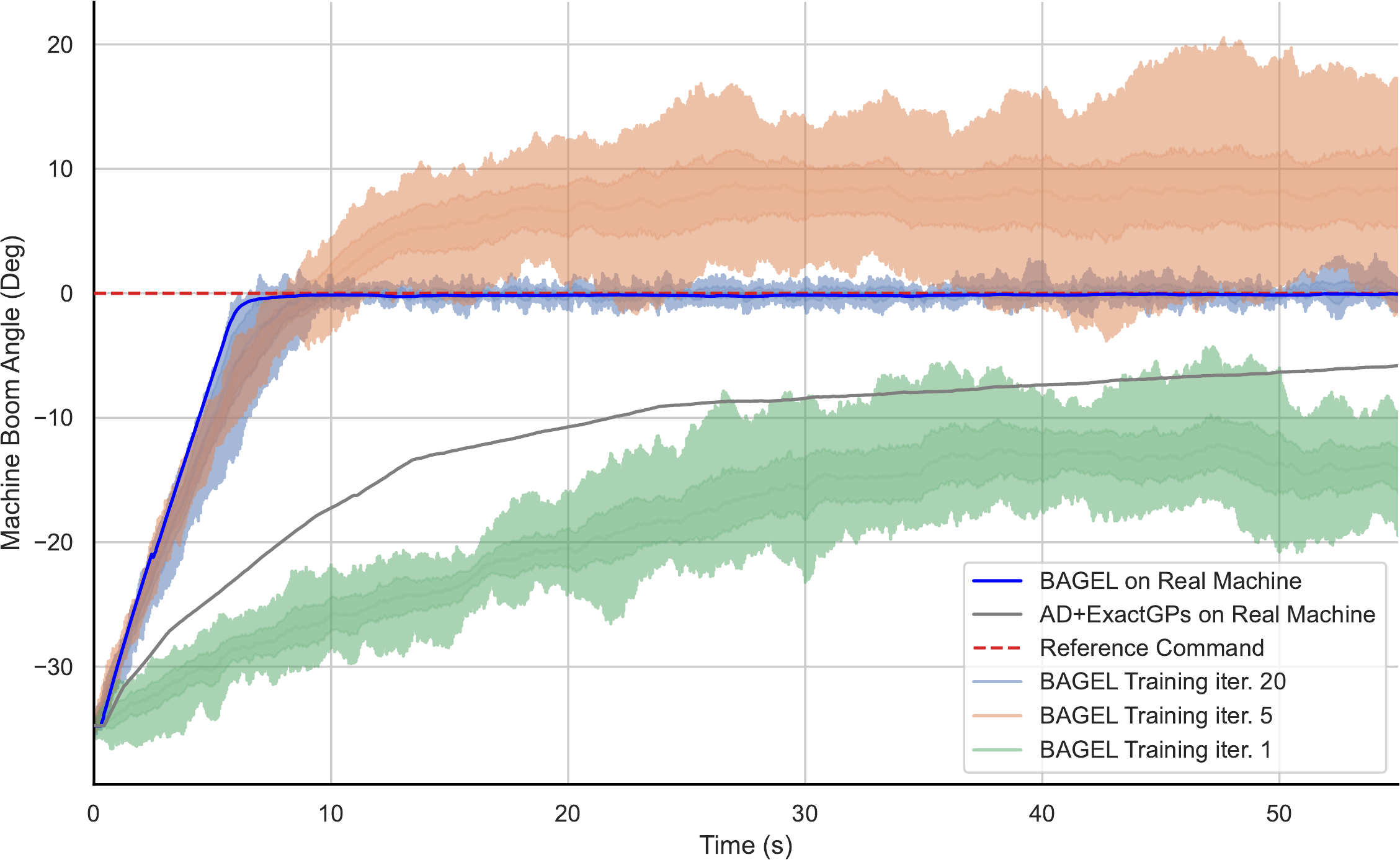}} 
  \caption{Real experiment results on a hydraulic boom using BAGEL. The ``imagined behavior" of the closed-loop system of NN+GPs are realized at different stages of training. Each shaded propagation contains the upper and lower bounds, 95\% confidence (darker) region, and mean of all predicted trajectories in a forward pass. BAGEL is trained for ${\raise.17ex\hbox{$\scriptstyle\sim$}}30s$ (20 iters) whereas AutoDiff+ExactGPs is given 5 hours for training (100 iters).}
  \label{fig:Traj1}
\end{figure}

Fig.~\ref{fig:Traj1} presents the results of applying the NN controller trained by BAGEL to the machine shown in Fig.~\ref{fig:avant}. This successfully drives the boom to the commanded position at $0^\circ$. The hydraulic actuator is fast reacting and has an intrinsic deadband, both of these features contribute to keeping the response steady during motion and without jitter at goal point. Because of the choice of state-dependent reward function (\ref{eqn_rew}), the solution resembles bang-bang control\cite{seyde2021bang}. Additional insights are also made possible by plotting the rollouts of BAGEL during early, middle, and late stages of training in Fig.~\ref{fig:Traj1}. Given the batch evaluation feature of the algorithm, we can see that the policy reduces the variance over all trajectories.

To showcase the performance gain of BAGEL as a result of using scalable GP querying \cite{gardner2018gpytorch}, we consider two other algorithms. In the first one, autodiff on exact GPs, we have modified the querying part in BAGEL-CPU-Seq to sample directly from cholesky decomposition methods. For autodiff on PILCO, a modified version of \cite{polymenakos2017safe} is used. As we have observed during the optimization process illustrated in Fig.~\ref{fig:ComputeTime}, one major issue that contributes to the drop of performance in both methods is the high demand of memory that causes constant swap between RAM and disk. To alleviate this issue to some extent, we let these methods train for several hours, yet both approaches are much slower in computation and the actuation of the policy (Fig.~\ref{fig:Traj1}) is quite slow for the machine. Moreover, the optimization benchmarks exclude the time it takes to learn the GP hyperparameters (\ref{eqn_loglikelihoodupdate}), around ${\raise.17ex\hbox{$\scriptstyle\sim$}}20s$ performed once for all algorithms, and the one-time caching operation required for BAGEL's fast predictions \cite{pleiss2018constant} takes ${\raise.17ex\hbox{$\scriptstyle\sim$}}0.6s$.

\begin{figure*}[!tp]
  \centering
  {\includegraphics[width=0.85\textwidth]{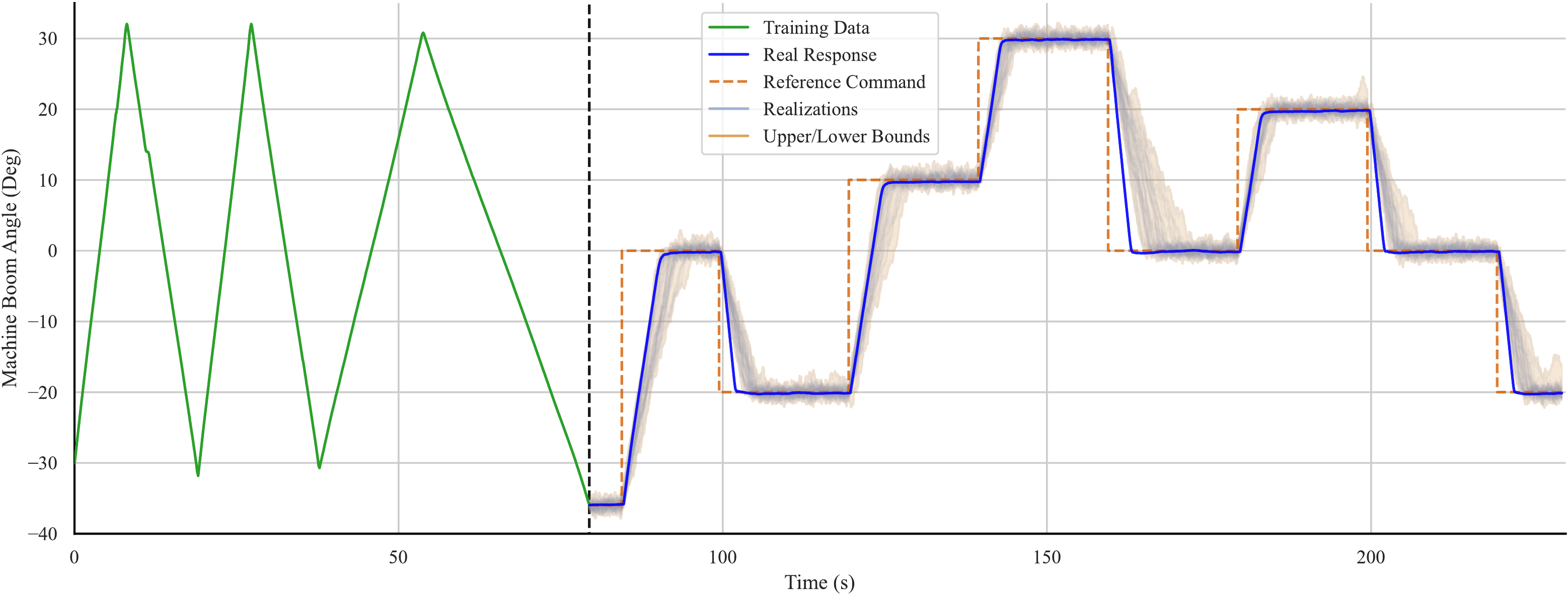}} 
  \caption{Real machine experiments with a goal-conditioned controller. In the first part ($t < 80s$), machine is controlled manually at two speeds to collect data. Then, a generalized goal-conditioned controller is trained using BAGEL and set to follow reference goals. The controller effectively drives all realizations of the sampled transitions to the target positions. Shaded region illustrates the bounds of realized trajectories.}
  \label{fig:Traj2} 
\end{figure*}

\subsection{Experiment 2: Goal-conditioned Policy} \label{sec:experiment2gc}

In our second experiment, we show the flexibility of our method in training a generalized, goal-conditioned policy $\pi_\theta(\bm{x}, \bm{x_g})$ that takes a varying goal state as additional input. Here, the reward function in (\ref{eqn_rew}) is now $r([\hat{\varphi}, \hat{\dot{\varphi}}], [\hat{\varphi}_g, \hat{\dot{\varphi}}_g])$. The batch of $\mathcal{S,G}$ for the training are sampled from a uniform distribution in their respective bounds from $\mathcal{D}$. This will result in the objective of maximizing rewards from any state to any goal in admissible space. Fig.~\ref{fig:Traj2} demonstrates the thorough process of collecting data by first manually controlling the machine, then sending step reference commands to the goal-conditioned controller to follow. The rollouts of the closed-loop system are drawn with thin lines, where all realizations are controlled by the policy.

\subsection{Scalability} \label{sec:scalability}
In Section \ref{sec:experiment1}, it was demonstrated that policy optimization becomes very demanding on memory with exact methods. This was expected to some extent since exact GP inference comes with large memory requirements. However, it is the integration of autodiff that demands more memory than what is normally required for GP inference to store gradients. Although BAGEL was well below the 4GB memory requirements of a laptop GPU during our experiments in Sections (\ref{sec:experiment1}-\ref{sec:experiment2gc}), we are interested in seeing its potential to scale to larger problems. Therefore, in this section we study the optimization speed and required memory with respect to three main hyperparameters that are the most important contributing factors to scaling: batch size $BS$, size of policy parameters $\bm{\theta}$, and horizon $H$. 

Fig.~\ref{fig:keops} summarizes the results of the analysis. Considering as a baseline the hyperparameters in our previous experiments (Sections \ref{sec:experiment1}-\ref{sec:experiment2gc}), we begin by scaling the number of trajectories processed in every batch by a factor of 10 and using a larger policy. As can be seen in Fig.~\ref{fig:keops}-(b), there is almost no drop in performance of BAGEL in the GPyTorch \cite{gardner2018gpytorch} implementation, as long as the memory demand for autodiff is provided. To scale beyond the GPU memory limits, there are a variety of methods \cite{margossian2019review} to reconstruct BAGEL into a memory-efficient implementation. One can process batches by breaking them into smaller batches or checkpoint gradients for larger policies \cite{chen2016training} at the expense of additional forward pass, in essence trading off more compute for less memory. For our benchmarks, we choose KeOps \cite{charlier2021kernel} which is easy to integrate with BAGEL and does not introduce additional hyperparameters. KeOps automatically builds memory-management on the kernel operations of BAGEL, including autodiff, allowing it to scale the succeeding configurations shown in Fig.~\ref{fig:keops} up to a horizon of 1K steps, larger policy, and batch size up to 10K. From the benchmarks and the structure of BAGEL in Algorithm~\ref{Algo:I}, it can be seen that scaling with respect to batch size and size of policy even after plugging in memory-efficiency is sublinear, whereas scaling with respect to H is near-linear since it increases the number of calls in the forward pass. The evaluations are average over 100 iterations, and omit the one-time compilation delay of KeOps functions. Note that a better utilization of memory in the last configuration in Fig.~\ref{fig:keops} is faster than the previous counterpart with smaller policy.


\section{Concluding Remarks} \label{sec:conclusion}

Not so long ago it seemed that inference and learning with Gaussian processes (GPs) was greatly impaired by computational constraints. However, recent algorithmic breakthroughs have not only overcome those challenges, but have accelerated inference by leveraging modern hardware. We re-examined controller learning using the latest techniques in GP prediction, optimization, and memory efficiency, and unified them into BAGEL. Throughout our experiments, we showcased how BAGEL can train policies for real machine control to track various reference trajectories in a timely manner, as well as its ability to scale up to larger problems and the flexibility to obtain generalized policies. 

Nevertheless, BAGEL is by no means flawless. The accuracy of model affects the quality of policy, and BAGEL naturally inherits the shortcomings of its underlying methods, such as limitations to scale to high-dimensional problems. There are many underlying components influencing the performance of BAGEL, so a projection to more dimensions, complex models and different tasks can only be made through problem-specific feasibility studies. Notwithstanding, BAGEL is a practical and flexible framework that can integrate well with a variety of learning techniques and extend to larger scale problems for future research.  

\section*{Acknowledgment}
The authors wish to thank Marcus Rösth and Pelle Gustafsson (HIAB), Mohammad M. Aref and Juho Vihonen (Cargotec) for their valuable feedback and discussion regarding heavy machines, Pascal Klink for insigths on memory efficiency, CSC IT Center for Science in Finland for computational resources, and Nikolay Serbenyuk, Nataliya Strokina, and Mehmet Killioglu for assisting with the machine experiments and data. This work has received funding from the European Union’s Horizon 2020 research and innovation programme Marie Skłodowska Curie under grant agreement No. 858101.

\bibliographystyle{ieeeconf}
\bibliography{references}

\begin{thebibliography}{10}
\providecommand{\url}[1]{#1}
\csname url@samestyle\endcsname
\providecommand{\newblock}{\relax}
\providecommand{\bibinfo}[2]{#2}
\providecommand{\BIBentrySTDinterwordspacing}{\spaceskip=0pt\relax}
\providecommand{\BIBentryALTinterwordstretchfactor}{4}
\providecommand{\BIBentryALTinterwordspacing}{\spaceskip=\fontdimen2\font plus
\BIBentryALTinterwordstretchfactor\fontdimen3\font minus
  \fontdimen4\font\relax}
\providecommand{\BIBforeignlanguage}[2]{{%
\expandafter\ifx\csname l@#1\endcsname\relax
\typeout{** WARNING: IEEEtran.bst: No hyphenation pattern has been}%
\typeout{** loaded for the language `#1'. Using the pattern for}%
\typeout{** the default language instead.}%
\else
\language=\csname l@#1\endcsname
\fi
#2}}
\providecommand{\BIBdecl}{\relax}
\BIBdecl

\bibitem{stooke2018accelerated}
A.~Stooke and P.~Abbeel, ``Accelerated methods for deep reinforcement
  learning,'' \emph{arXiv preprint arXiv:1803.02811}, 2018.

\bibitem{mnih2016asynchronous}
V.~Mnih, A.~P. Badia, M.~Mirza, A.~Graves, T.~Lillicrap, T.~Harley, D.~Silver,
  and K.~Kavukcuoglu, ``Asynchronous methods for deep reinforcement learning,''
  in \emph{International conference on machine learning}.\hskip 1em plus 0.5em
  minus 0.4em\relax PMLR, 2016, pp. 1928--1937.

\bibitem{schulman2017proximal}
J.~Schulman, F.~Wolski, P.~Dhariwal, A.~Radford, and O.~Klimov, ``Proximal
  policy optimization algorithms,'' \emph{arXiv preprint arXiv:1707.06347},
  2017.

\bibitem{chatzilygeroudis2017black}
K.~Chatzilygeroudis, R.~Rama, R.~Kaushik, D.~Goepp, V.~Vassiliades, and J.-B.
  Mouret, ``Black-box data-efficient policy search for robotics,'' in
  \emph{2017 IEEE/RSJ International Conference on Intelligent Robots and
  Systems (IROS)}.\hskip 1em plus 0.5em minus 0.4em\relax IEEE, 2017, pp.
  51--58.

\bibitem{langaaker2018cautious}
H.-A. Lang{\aa}ker, ``Cautious mpc-based control with machine learning,''
  Master's thesis, NTNU, 2018.

\bibitem{andersson2021reinforcement}
J.~Andersson, K.~Bodin, D.~Lindmark, M.~Servin, and E.~Wallin, ``Reinforcement
  learning control of a forestry crane manipulator,'' in \emph{2021 IEEE/RSJ
  International Conference on Intelligent Robots and Systems (IROS)}.\hskip 1em
  plus 0.5em minus 0.4em\relax IEEE, 2021, pp. 2121--2126.

\bibitem{lindmark2018computational}
D.~M. Lindmark and M.~Servin, ``Computational exploration of robotic rock
  loading,'' \emph{Robotics and Autonomous Systems}, vol. 106, pp. 117--129,
  2018.

\bibitem{muratore2021robot}
F.~Muratore, F.~Ramos, G.~Turk, W.~Yu, M.~Gienger, and J.~Peters, ``Robot
  learning from randomized simulations: A review,'' \emph{arXiv preprint
  arXiv:2111.00956}, 2021.

\bibitem{klink2021model}
P.~Klink, ``Model-based reinforcement learning from pilco to pets,'' in
  \emph{Reinforcement Learning Algorithms: Analysis and Applications}.\hskip
  1em plus 0.5em minus 0.4em\relax Springer, 2021, pp. 165--175.

\bibitem{moerland2020model}
T.~M. Moerland, J.~Broekens, and C.~M. Jonker, ``Model-based reinforcement
  learning: A survey,'' \emph{arXiv preprint arXiv:2006.16712}, 2020.

\bibitem{janner2019trust}
M.~Janner, J.~Fu, M.~Zhang, and S.~Levine, ``When to trust your model:
  Model-based policy optimization,'' \emph{Advances in Neural Information
  Processing Systems}, vol.~32, 2019.

\bibitem{polydoros2017survey}
A.~S. Polydoros and L.~Nalpantidis, ``Survey of model-based reinforcement
  learning: Applications on robotics,'' \emph{Journal of Intelligent \& Robotic
  Systems}, vol.~86, no.~2, pp. 153--173, 2017.

\bibitem{williams2006gaussian}
C.~K. Williams and C.~E. Rasmussen, \emph{Gaussian processes for machine
  learning}.\hskip 1em plus 0.5em minus 0.4em\relax MIT press Cambridge, MA,
  2006, vol.~2, no.~3.

\bibitem{deisenroth2011pilco}
M.~Deisenroth and C.~E. Rasmussen, ``Pilco: A model-based and data-efficient
  approach to policy search,'' in \emph{Proceedings of the 28th International
  Conference on machine learning (ICML-11)}.\hskip 1em plus 0.5em minus
  0.4em\relax Citeseer, 2011, pp. 465--472.

\bibitem{deisenroth2014multi}
M.~P. Deisenroth, P.~Englert, J.~Peters, and D.~Fox, ``Multi-task policy search
  for robotics,'' in \emph{2014 IEEE International Conference on Robotics and
  Automation (ICRA)}.\hskip 1em plus 0.5em minus 0.4em\relax IEEE, 2014, pp.
  3876--3881.

\bibitem{calandra2014experimental}
R.~Calandra, A.~Seyfarth, J.~Peters, and M.~P. Deisenroth, ``An experimental
  comparison of bayesian optimization for bipedal locomotion,'' in \emph{2014
  IEEE International Conference on Robotics and Automation (ICRA)}.\hskip 1em
  plus 0.5em minus 0.4em\relax IEEE, 2014, pp. 1951--1958.

\bibitem{delgado2020sample}
J.~A. Delgado-Guerrero, A.~Colom{\'e}, and C.~Torras, ``Sample-efficient robot
  motion learning using gaussian process latent variable models,'' in
  \emph{2020 IEEE International Conference on Robotics and Automation
  (ICRA)}.\hskip 1em plus 0.5em minus 0.4em\relax IEEE, 2020, pp. 314--320.

\bibitem{deisenroth2013gaussian}
M.~P. Deisenroth, D.~Fox, and C.~E. Rasmussen, ``Gaussian processes for
  data-efficient learning in robotics and control,'' \emph{IEEE transactions on
  pattern analysis and machine intelligence}, vol.~37, no.~2, pp. 408--423,
  2013.

\bibitem{pleiss2018constant}
G.~Pleiss, J.~Gardner, K.~Weinberger, and A.~G. Wilson, ``Constant-time
  predictive distributions for gaussian processes,'' in \emph{International
  Conference on Machine Learning}.\hskip 1em plus 0.5em minus 0.4em\relax PMLR,
  2018, pp. 4114--4123.

\bibitem{gardner2018gpytorch}
J.~Gardner, G.~Pleiss, K.~Q. Weinberger, D.~Bindel, and A.~G. Wilson,
  ``Gpytorch: Blackbox matrix-matrix gaussian process inference with gpu
  acceleration,'' \emph{Advances in neural information processing systems},
  vol.~31, 2018.

\bibitem{chatzilygeroudis2019survey}
K.~Chatzilygeroudis, V.~Vassiliades, F.~Stulp, S.~Calinon, and J.-B. Mouret,
  ``A survey on policy search algorithms for learning robot controllers in a
  handful of trials,'' \emph{IEEE Transactions on Robotics}, vol.~36, no.~2,
  pp. 328--347, 2019.

\bibitem{backman2021continuous}
S.~Backman, D.~Lindmark, K.~Bodin, M.~Servin, J.~M{\"o}rk, and H.~L{\"o}fgren,
  ``Continuous control of an underground loader using deep reinforcement
  learning,'' \emph{Machines}, vol.~9, no.~10, p. 216, 2021.

\bibitem{kurinov2020automated}
I.~Kurinov, G.~Orzechowski, P.~H{\"a}m{\"a}l{\"a}inen, and A.~Mikkola,
  ``Automated excavator based on reinforcement learning and multibody system
  dynamics,'' \emph{IEEE Access}, vol.~8, pp. 213\,998--214\,006, 2020.

\bibitem{berglund2021controlling}
D.~Berglund and N.~Larsson, ``Controlling a hydraulic system using
  reinforcement learning: Implementation and validation of a dqn-agent on a
  hydraulic multi-chamber cylinder system,'' Master's thesis, Link{\"o}ping
  University, 2021.

\bibitem{yang2021neural}
W.~Yang, N.~Strokina, N.~Serbenyuk, J.~Pajarinen, R.~Ghabcheloo, J.~Vihonen,
  M.~M. Aref, and J.-K. K{\"a}m{\"a}r{\"a}inen, ``Neural network controller for
  autonomous pile loading revised,'' in \emph{2021 IEEE International
  Conference on Robotics and Automation (ICRA)}.\hskip 1em plus 0.5em minus
  0.4em\relax IEEE, 2021, pp. 2198--2204.

\bibitem{backas2019nonlinear}
J.~Backas and R.~Ghabcheloo, ``Nonlinear model predictive energy management of
  hydrostatic drive transmissions,'' \emph{Proceedings of the Institution of
  Mechanical Engineers, Part I: Journal of Systems and Control Engineering},
  vol. 233, no.~3, pp. 335--347, 2019.

\bibitem{kamthe2018data}
S.~Kamthe and M.~Deisenroth, ``Data-efficient reinforcement learning with
  probabilistic model predictive control,'' in \emph{International conference
  on artificial intelligence and statistics}.\hskip 1em plus 0.5em minus
  0.4em\relax PMLR, 2018, pp. 1701--1710.

\bibitem{snelson2007local}
E.~Snelson and Z.~Ghahramani, ``Local and global sparse gaussian process
  approximations,'' in \emph{Artificial Intelligence and Statistics}.\hskip 1em
  plus 0.5em minus 0.4em\relax PMLR, 2007, pp. 524--531.

\bibitem{hewing2019cautious}
L.~Hewing, J.~Kabzan, and M.~N. Zeilinger, ``Cautious model predictive control
  using gaussian process regression,'' \emph{IEEE Transactions on Control
  Systems Technology}, vol.~28, no.~6, pp. 2736--2743, 2019.

\bibitem{snelson2006sparse}
E.~Snelson and Z.~Ghahramani, ``Sparse gaussian processes using
  pseudo-inputs,'' \emph{Advances in neural information processing systems},
  vol.~18, p. 1257, 2006.

\bibitem{wilson2021pathwise}
J.~T. Wilson, V.~Borovitskiy, A.~Terenin, P.~Mostowsky, and M.~P. Deisenroth,
  ``Pathwise conditioning of gaussian processes,'' \emph{Journal of Machine
  Learning Research}, vol.~22, no. 105, pp. 1--47, 2021.

\bibitem{wilson2015kernel}
A.~Wilson and H.~Nickisch, ``Kernel interpolation for scalable structured
  gaussian processes (kiss-gp),'' in \emph{International Conference on Machine
  Learning}.\hskip 1em plus 0.5em minus 0.4em\relax PMLR, 2015, pp. 1775--1784.

\bibitem{charlier2021kernel}
B.~Charlier, J.~Feydy, J.~Glaun{\`e}s, F.-D. Collin, and G.~Durif, ``Kernel
  operations on the gpu, with autodiff, without memory overflows,''
  \emph{Journal of Machine Learning Research}, vol.~22, no.~74, pp. 1--6, 2021.

\bibitem{rudi2017falkon}
A.~Rudi, L.~Carratino, and L.~Rosasco, ``Falkon: An optimal large scale kernel
  method,'' \emph{Advances in neural information processing systems}, vol.~30,
  2017.

\bibitem{wang2019exact}
K.~Wang, G.~Pleiss, J.~Gardner, S.~Tyree, K.~Q. Weinberger, and A.~G. Wilson,
  ``Exact gaussian processes on a million data points,'' \emph{Advances in
  Neural Information Processing Systems}, vol.~32, pp. 14\,648--14\,659, 2019.

\bibitem{meanti2020kernel}
G.~Meanti, L.~Carratino, L.~Rosasco, and A.~Rudi, ``Kernel methods through the
  roof: handling billions of points efficiently,'' \emph{Advances in Neural
  Information Processing Systems}, vol.~33, pp. 14\,410--14\,422, 2020.

\bibitem{umlauft2017learning}
J.~Umlauft, A.~Lederer, and S.~Hirche, ``Learning stable gaussian process state
  space models,'' in \emph{2017 American Control Conference (ACC)}.\hskip 1em
  plus 0.5em minus 0.4em\relax IEEE, 2017, pp. 1499--1504.

\bibitem{he2015delving}
K.~He, X.~Zhang, S.~Ren, and J.~Sun, ``Delving deep into rectifiers: Surpassing
  human-level performance on imagenet classification,'' in \emph{Proceedings of
  the IEEE international conference on computer vision}, 2015, pp. 1026--1034.

\bibitem{paszke2019pytorch}
A.~Paszke, S.~Gross, F.~Massa, A.~Lerer, J.~Bradbury, G.~Chanan, T.~Killeen,
  Z.~Lin, N.~Gimelshein, L.~Antiga \emph{et~al.}, ``Pytorch: An imperative
  style, high-performance deep learning library,'' \emph{Advances in neural
  information processing systems}, vol.~32, pp. 8026--8037, 2019.

\bibitem{kingma2014adam}
D.~P. Kingma and J.~Ba, ``Adam: A method for stochastic optimization,''
  \emph{arXiv preprint arXiv:1412.6980}, 2014.

\bibitem{seyde2021bang}
T.~Seyde, I.~Gilitschenski, W.~Schwarting, B.~Stellato, M.~Riedmiller,
  M.~Wulfmeier, and D.~Rus, ``Is bang-bang control all you need? solving
  continuous control with bernoulli policies,'' \emph{Advances in Neural
  Information Processing Systems}, vol.~34, 2021.

\bibitem{polymenakos2017safe}
K.~Polymenakos, A.~Abate, and S.~Roberts, ``Safe policy search with gaussian
  process models,'' \emph{arXiv preprint arXiv:1712.05556}, 2017.

\bibitem{margossian2019review}
C.~C. Margossian, ``A review of automatic differentiation and its efficient
  implementation,'' \emph{Wiley interdisciplinary reviews: data mining and
  knowledge discovery}, vol.~9, no.~4, p. e1305, 2019.

\bibitem{chen2016training}
T.~Chen, B.~Xu, C.~Zhang, and C.~Guestrin, ``Training deep nets with sublinear
  memory cost,'' \emph{arXiv preprint arXiv:1604.06174}, 2016.

\end{thebibliography}

\end{document}